# FAQ-Gen: An automated system to generate domain-specific FAQs to aid content comprehension

Sahil Kale[*1], Gautam Khaire[1], Jay Patankar[1]
[1]Pune Institute of Computer Technology, Pune, India
*Corresponding author: sahilrkale05@gmail.com

*Abstract*

Frequently Asked Questions (FAQs) refer to the most common inquiries about specific content. They serve as content comprehension aids by simplifying topics and enhancing understanding through succinct presentation of information. In this paper, we address FAQ generation as a well-defined Natural Language Processing task through the development of an end-to-end system leveraging text-to-text transformation models. We present a literature review covering traditional question-answering systems, highlighting their limitations when applied directly to the FAQ generation task. We propose a system capable of building FAQs from textual content tailored to specific domains, enhancing their accuracy and relevance. We utilise self-curated algorithms to obtain an optimal representation of information to be provided as input and also to rank the question-answer pairs to maximise human comprehension. Qualitative human evaluation showcases the generated FAQs as well-constructed and readable while also utilising domain-specific constructs to highlight domain-based nuances and jargon in the original content.

## 1. INTRODUCTION

Frequently Asked Questions (FAQs) for a particular subject matter are defined as a ranked collection of the most commonly asked questions about the topic accompanied by the corresponding answers (Raazaghi 2015). FAQs can help address several content-comprehension problems by reducing the associated understanding time, answering common queries and doubts efficiently, and quickly resolving pain points and gaps in understanding the given content. Thus, to leverage the impact of these powerful text comprehension aids, referred to as FAQs through website terminology on Wikipedia[1], we propose a complete automated system that can analyse important information from given textual contexts and produce high-quality, clear, and concise FAQs with easy-to-understand answers through this paper. Figure 1 shows an example of high-quality FAQs (shown on the right) that encapsulate the entire paragraph context with well-formed and intuitive answers and also contrasts them with low-quality counterparts (shown on the left) that focus only on one part of the entire paragraph with answers that are neither elaborate nor human-readable.

The problems of question generation (QG) and question-answering (QA) in the field of Natural Language Processing (NLP) have received significant interest in recent years in industrial as well as academic contexts (Zhou et al. 2017; Zhao et al. 2018; Chan and Fan 2019). However, even though existing question-generation models employ powerful neural networks to generate highly relevant and comprehensive questions, pipelining and linking consequent question-answering models that automatically produce straightforward, easy-to-read answers to these questions remains a difficult task. Furthermore, most existing approaches utilise answer-aware question-generation techniques, which can limit and hinder the quality and completeness of the questions being generated from a given context (Du and Cardie 2018; Hu et al. 2018; Liu et al. 2020). Also, domain-specific question generation and answering capability are less explored in current systems. These drawbacks of utilising existing models for the tasks of FAQ creation and answering bring to light the need for a complete, end-to-end system for the automated generation and management of FAQs.

| **Paragraph:** | **Paragraph:** |
|---|---|
| Entanglement, the sole hallmark of quantum mechanics, describes a peculiar connection between particles that enables them to instantaneously influence each other's states, regardless of distance. This phenomenon has been experimentally verified and forms the basis for quantum technologies like quantum cryptography and quantum computing. While quantum mechanics has achieved remarkable success in explaining microscopic phenomena, its counterintuitive aspects, such as uncertainty and non-locality, continue to intrigue scientists and challenge our conventional understanding of reality. | Entanglement, the sole hallmark of quantum mechanics, describes a peculiar connection between particles that enables them to instantaneously influence each other's states, regardless of distance. This phenomenon has been experimentally verified and forms the basis for quantum technologies like quantum cryptography and quantum computing. While quantum mechanics has achieved remarkable success in explaining microscopic phenomena, its counterintuitive aspects, such as uncertainty and non-locality, continue to intrigue scientists and challenge our conventional understanding of reality. |
| **FAQs:** | **FAQs:** |
| - What is the sole hallmark of quantum mechanics? | - What is the sole hallmark of quantum mechanics? |
| *"entanglement"* | *"Entanglement is the sole hallmark of quantum mechanics."* |
| - What does entanglement describe? | - What technologies does entanglement form the basis for? |
| *"connection between particles"* | *"Entanglement forms the basis for quantum technologies like quantum cryptography and quantum computing"* |
| - What does the connection between particles enable? | - What are the counterintuitive aspects of quantum mechanics? |
| *"influence each other's states"* | *"Quantum mechanics has certain counterintuitive aspects such as uncertainty and non-locality"* |

FIGURE 1. EXAMPLE OF LOW-QUALITY (LEFT) AND HIGH-QUALITY (RIGHT) FAQS.

---

[1] https://en.wikipedia.org/wiki/FAQ

The complete process flow of the automated FAQ generation and management task can be divided into multiple subtasks, which include identifying the domain of the input content so as to utilise domain-specific vocabulary and constructs in the further process, generating questions from given textual content according to the identified domain, finding out answer keywords/keyphrases for the curated questions, elaborating the answer keywords/keyphrases to make them human-readable and finally ranking question-answer pairs according to their relevance and usefulness for end users.

The contribution of this paper can be summarised as follows:

- We identify and analyse the shortcomings of existing approaches and systems for automated FAQ generation and put forth areas where improvements can be made.

- We propose and describe a simple but effective architecture for an end-to-end, domain-specific FAQ generation system tackling the subtasks of domain identification, question generation, question-answering, answer elaboration, and question-answer pair ranking.

- We demonstrate and analyse the results of the FAQ generation system on data from various sources and demonstrate the innovativeness of our system and its applicability for large-scale implementations and future development.

The rest of this paper is structured as follows. In Section 2, we analyse and review related work and similar systems for FAQ generation while also discussing the base model used in our implementation. In Section 3, we describe in depth the architecture and workings of our system along with the complete process flow. Section 4 discusses the innovativeness of our approach and optimisation techniques used, and Section 5 provides results and inference. In Section 6, we conclude the paper and discuss future improvements.

## 2. LITERATURE SURVEY

### 2.1. Review of Related Work

The major subtasks of the FAQ generation system, namely question-generation and question-answering, have received significant attention due to their wide-ranging applications in machine reading comprehension, chatbots, and automated learning systems (Kumar et al. 2019). Closed-domain question-generation systems for specific fields like education (Kunichika et al. 2004) and medicine (Shen et al. 2020) produce valuable results in their respective domains, the idea of which is expanded in our system by producing domain-specific NLP models in the pipeline.

Answer assessment strategies suggested by Das et al. (2021) have been taken into note to build an answer elaboration model that makes answers to questions to the point and easy to understand. The datasets SQuAD (Rajpurkar et al. 2016), NewsQA (Trischler et al. 2018),

TriviaQA (Joshi et al. 2017), and NarrativeQA (Kočiský et al. 2017) containing question-answer pairs were analysed for their suitability in training models for our system. However, due to a lack of diversity in topics and insufficient information, the need to generate new datasets was identified. Furthermore, a review of current open-source FAQ generation tools and packages like Questgen AI[2] and Question Generator[3] shows that these systems often struggle with complex textual information and diverse or short contexts as they rely on domain-specific training data and manual rule creation.

Table 1 summarises the most relevant research work in building a comprehensive, end-to-end FAQ generation system, with a focus on work that can contribute to implementing at least one subtask of the FAQ generation system described in Section 1, namely domain identification, question generation, question-answering, answer elaboration, and question-answer pair ranking. Drawbacks mentioned in the resources or identified through suitability analysis have also been included in the table to highlight the difficulty of using existing techniques and systems directly for our requirements. Our system aims to overcome these drawbacks to generate high-quality, clear, and concise FAQs in a domain-aware manner.

| Paper | Relevance for FAQ system | Drawbacks Identified |
|---|---|---|
| Lu and Lu (2021) | Explores techniques for automatic question generation useful for FAQ generation subtasks, highlighting rule-based and supervised approaches | Lack of evaluation metrics can degrade the performance of question generation models, and lack of training data for FAQ generation tasks |
| Zhang et al. (2023) | Proposes an approach for multiple session-based question answering using attention-based mechanisms supported with historical contexts, useful for text-based FAQ system | Limited testing for question-generation and answering tasks and a lack of domain-specific knowledge for question-answering |
| Zhang and Bansal (2019) | Proposes semantics-enhanced rewards for generated questions, ensuring they adhere to original content and are easy to answer, a requirement of FAQs | Requires downstream question paraphrasing and answering tasks, adding complexity to building a complete pipeline as required for an FAQ generation system |
| Roemmele et al. (2021) | Integrates question | Limited evaluation on specific |

---

[2] https://github.com/ramsrigouthamg/Questgen.ai
[3] https://github.com/AMontgomerie/question_generator

| | answering with question generation to produce QA items in multi-paragraph content simulating FAQ generation behaviour | datasets, requires adaptation for new domains and document styles |
|---|---|---|
| Du et al. (2017) | Introduces an attention-based model for generating high-quality, diverse questions directly from text passages, satisfying the requirements of question quality for FAQs | Achieves best performance for the question-generation task, but failure to set the difficulty level and type of questions hinders its applicability for an FAQ system pipeline |

TABLE 1. FAQ GENERATION SYSTEM - RELEVANT WORK AND DRAWBACKS

## 2.2. Base Model Selection

The process of transfer learning has proven to be tremendously powerful in the field of Natural Language Processing owing to its ability to fine-tune models for downstream tasks while preserving the pre-training on a data-rich task (Ruder et al. 2019). The Bidirectional Encoder Representations from Transformers (BERT) (Devlin et al. 2019) and Text-to-Text Transfer Transformer (T5) (Raffel et al. 2023) models are popular 'base models' as they are pre-trained on a large corpus of text and can be fine-tuned for a variety of different tasks, as seen through the applications compiled by Alyafeai et al. (2020).

The BERT model is a state-of-the-art natural language processing model with a powerful ability to understand context in text data. Its bidirectional nature allows it to capture dependencies and relationships effectively (Devlin et al. 2019). This strength of understanding contextual nuances within sentences makes it proficient in the question-answering subtask of the automated FAQ generation system. However, as the BERT model operates primarily in a pre-trained context-independent manner, it falls short in question generation and answer elaboration subtasks, which can potentially lead to less diverse and contextually rich outputs.

The T5 model, being an encoder-decoder model, has the ability to convert all NLP problems into a text-to-text transformation format, a feature immensely important for the question-generation, question-answering, and answer elaboration subtasks of the FAQ generation system. The abilities of the T5 model to be used for a variety of tasks by simply prepending a different prefix to the input corresponding to each task and to be trained for text-to-text mappings by providing supervised examples of input sequences and corresponding target sequences (Raffel et al. 2023) make it an ideal choice as the base model to be fine-tuned for a variety of FAQ system subtasks.

## 3. METHODOLOGY

## 3.1. Approach Overview

The approach proposed in this paper adopts a modular style to achieve the goal of FAQ generation, as described in Figure 2, while satisfying all the required qualities of good, well-developed FAQs. To ensure appropriate and sufficient processing at each stage, the entire workflow of the FAQ generation task is segmented into six major and elaborate subtasks, referred to as steps henceforth. An example of the terminology used in the FAQ generation steps is shown in Figure 3. The steps for the entire process can be enumerated as follows:

(1) Text Extraction and Chunking

(2) Domain Identification

(3) Question Generation

(4) Answer Keyword/Keyphrase Extraction

(5) Answer Completion and Elaboration

(6) FAQ Compilation and Ranking

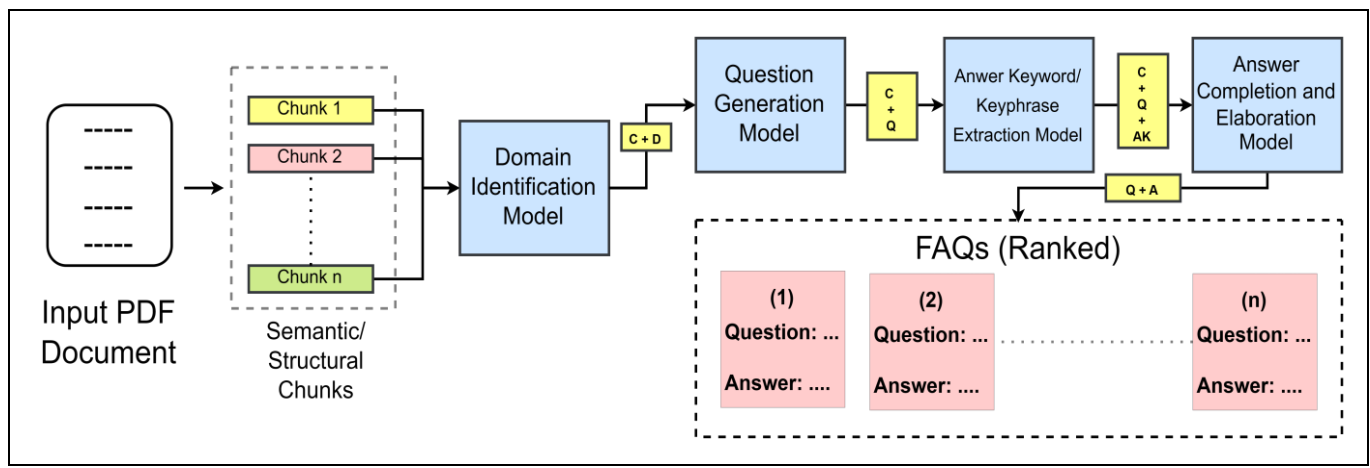

FIGURE 2. ARCHITECTURE OF THE FAQ GENERATION SYSTEM. (HERE, C = CONTEXT, D = DOMAIN, Q = QUESTION, AK = ANSWER KEYWORD/KEYPHRASE, A = COMPLETE ANSWER)

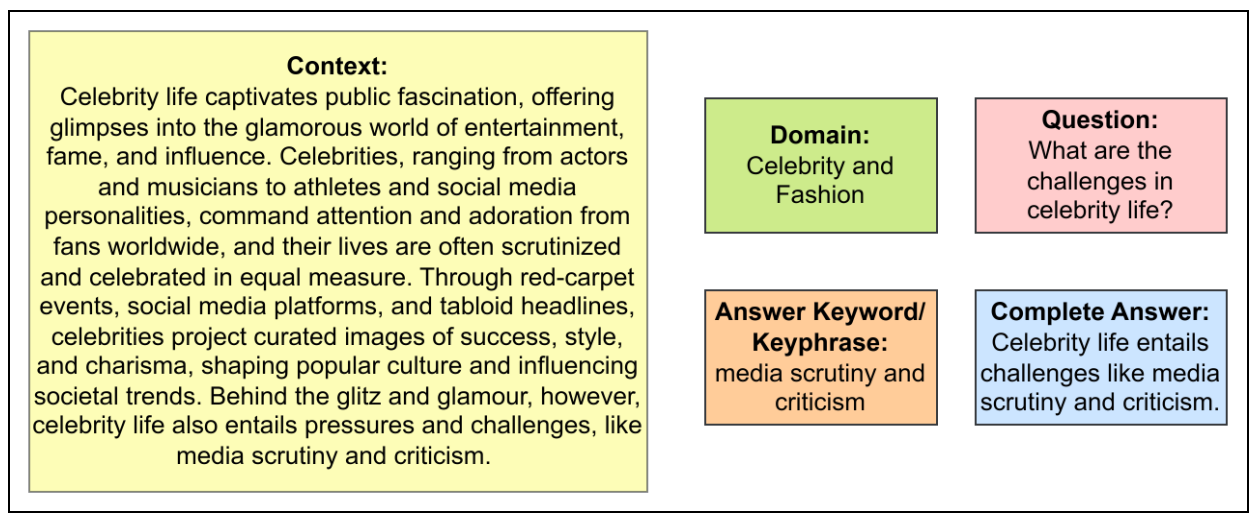

FIGURE 3. TERMINOLOGY USED IN THE FAQ GENERATION PROCESS WITH AN EXAMPLE

These steps are executed for each given input document to the system, ensuring that every document undergoes a structured and systematic progression through multiple processing stages. A brief discussion of every step is provided in the following sections.

### 3.1.1. Text Extraction and Chunking

The proposed FAQ generation system accepts a digital PDF document as input, owing to the versatility and portability of the file storage format. The first step involves the extraction of text, currently only in the English language, from the input document using common text-extraction tools like Py-PDF[4] and optical character recognition (OCR) (Kay 2007) models from the given document.

After extraction, in order to maintain the comprehensibility and completeness of FAQs generated from the document, the text is segmented into smaller units known as 'chunks,' each containing a distinct portion of the overall content. Within each chunk, the specific text under consideration is referred to as the 'context.' While both chunk and context relate to sections of the text, chunks represent the structural divisions of the input document, whereas contexts emphasise the textual content within each chunk. This context represents a coherent set of sentences serving as the primary corpus of information for subsequent steps, which generate question-answer pairs relevant to the context. Building chunks, i.e., chunking the input document, can be achieved in two ways as follows:

- **Semantic partitioning:** After textual content is obtained from the document, a sequence of sentences that elaborates a specific concept, process, or entity within the document is taken as one context by focusing on semantic relationships. For instance, information organised on Wikipedia can be semantically partitioned, giving rise to a structure similar to the sections on each Wiki page. The advantage of semantic partitioning lies in the fact that it ensures all related information occurs in a single context without putting restrictions on the size of the context. However, the possible downside lies in the worst case, where contexts of vastly disparate lengths are generated, presenting the textual information in a biased way to the models in subsequent steps.

- **Structural partitioning:** Structural partitioning restricts the size of each context by putting limits on the number of words or sentences in each context. Structural partitioning can generate contexts of roughly the same size. The fixed size not only ensures an unbiased representation of original information but also makes downstream tasks more systematic and predictable, as it is possible to make better decisions regarding the compilation and presentation of the results when the context size is known beforehand.

In structural partitioning, a crucial observation arises: when dividing a text consisting of $n$

---

[4] https://github.com/py-pdf/pypdf

words into chunks of size *m*, where m represents the desired chunk size, the text is partitioned into *n//m* chunks of size *m*, with an additional chunk of size *n%m*. Here, "//" denotes floor division (quotient of the division) and "%" denotes modulo (remainder of the division). This approach minimises the overall disparity in chunk sizes, especially for larger text sizes. For instance, structural partitioning would divide a text of 50,023 words into 200 chunks of the same size (~250 words), with only the last chunk of size 23 words being smaller than others. This would still outperform semantic chunking in terms of consistency of the size of chunks, as it would not have resulted in 200 chunks of approximately the same size.

Thus, in the context of our system, we use the length of the textual content to partition the text structurally into chunks. The chunk size, i.e., the number of words in the chunk context, is set to approximately 250 words, as it represents the size of a typical piece of self-reliant information. The partitioning logic chooses the first sentence where the total word count exceeds 250, as the last sentence for the chunk under construction, ensuring chunks do not end abruptly without completing the original sentence. The context from each chunk is then passed sequentially through steps 2-5, and the outputs obtained after the processing of all chunks is completed are compiled and organised in step 6.

### 3.1.2. Domain Identification

This step focuses on determining the domain to which the input text in a given chunk, i.e., the context, belongs. A domain here refers to a subject encompassing entities and events related to a specific theme, such as finance, adventure, or science and technology. A pre-trained text classification model (Antypas et al. 2022) is employed to map each context to the most relevant domain, a list of which can be found in Table 2. This pre-trained text classification model is based on a TimeLMs language model trained on around 124 million tweets from January 2018 to December 2021 and fine-tuned for the multi-label topic classification task on a corpus of 11,267 tweets. It has an F1 score of 0.588 and a Jaccard score of 0.676 for multi-label classification.

| Arts and Culture | Fitness and Health | News and Social Concern |
|---|---|---|
| Business and Entrepreneurs | Food and Dining | Science and Technology |
| Celebrity and Fashion | Gaming | Sports |
| Diaries and Daily Life | Learning and Educational | Travel and Adventure |
| Family and Relationships | Literature | Youth and Student Life |
| Film, TV and Video | Music | |

TABLE 2. A COMPLETE LIST OF DOMAINS USED IN THE FAQ GENERATION SYSTEM

The importance of the domain identification step lies in its immediate impact on the subsequent question-generation step, which is the only generative step in the FAQ generation process, as all the subsequent downstream steps involve cognition of existing knowledge from the questions and context. To generate nuanced, domain-specific FAQs, it is immensely important that the question-generation step accurately captures the vocabulary, tone, intricacy, and specificity in language constructs of the domain that the context belongs to, thereby making the domain identification step, which feeds the next step with the identified domain, a significant one.

### 3.1.3. Question Generation

In this phase, a set of unique and coherent questions is generated based on the provided context and its associated domain for each chunk. To aid this step, the approach proposed here uses domain-specific question-generation models for each domain, i.e., separate instances of T5 models fine-tuned to generate a list of questions based on the context for a dedicated domain. Using domain-specific datasets described in Section 3.2.1, we ensure that each question-generation model has an in-depth understanding of the nuanced language, terminology, and context inherent to that particular domain.

An open-source tool called SimpleT5[5] was utilised to fine-tune the T5 model for the required task. Hyperparameters such as the maximum token lengths for both source and target text were adjusted, with training comprising ten epochs and a batch size of one, owing to the large size of inputs. GPU acceleration was used to enhance the efficiency of the training procedure. After fine-tuning, the model is able to produce a comprehensive list of questions while maintaining a consistent output of approximately 4-5 questions for each provided context. This proficiency is attributed to the training dataset, which has been tailored and augmented as per the steps described in Section 3.2.1, which not only instructs the model on question generation logic but also ensures a comprehensive coverage of all the key points within the given context by guiding the model on the appropriate distribution of questions.

Each question-generation model tailored for a specific domain is thus able to craft WH-type, comparison-based, and cardinal questions specifically related to the original context. For instance, for a context identified to lie in the 'Science and Technology' domain, the generated questions resonate with the intricate details and unique jargon prevalent in the chosen field of interest, i.e. science. Each question, along with its associated context, is then passed downstream to the next step in the process.

### 3.1.4. Answer Keyword/Keyphrase Extraction

This step involves the use of a question-answering model to extract or generate an answer keyword or keyphrase for each question from its associated context. As the given task does not require domain-specific generative constructs, a single model is built for this step by using T5 as a base model and fine-tuning it with a custom dataset described in Section 3.2.2 for the given task. Here, too, the SimpleT5 tool was employed to conduct fine-tuning of the T5 model. The fine-tuning process included four epochs with a batch size of one and setting hyperparameters of maximum token lengths for both source and target text required in the transformation. A GPU was used to expedite the procedure.

---

[5] https://github.com/Shivanandroy/simpleT5

After the fine-tuning process, this model possesses the capability to identify a single keyword or a set of keywords (keyphrase) that can serve as the answer to a given question within the specified context. In order to elaborate on the identified answer keywords/keyphrases and form complete human-readable answers, each question, along with its associated answer keywords/keyphrase and context, is provided as input for the subsequent step.

This step is introduced in the FAQ generation process to overcome the drawbacks of direct question-answering using large language models, which generally focus only on the question and may ignore context, leading to incomplete and partial answers. By generating specific answer keywords or keyphrases for each question, we avoid generating generic or repetitive responses that lack depth while also ensuring that we can accurately pinpoint the source of answers for each FAQ from the original context to maximise transparency.

### 3.1.5. Answer Completion and Elaboration

The answer keywords or keyphrases generated in the previous step are further refined and elaborated on in this step by incorporating surrounding text from the context, ensuring the production of a grammatically correct, complete, and readable answer to each question. An answer-completion model is employed to accomplish this step of answer completion, which generates complete sentences that constitute appropriate answers within the given context for each question.

The text-to-text transformation required in this model is also achieved by fine-tuning the T5 base model with a custom dataset described in Section 3.2.2 using the SimpleT5 tool. The maximum token lengths for both source and target text were adjusted, with five epochs having a batch size of one used for the fine-tuning process accelerated by the use of a GPU.

The answer completion with elaboration step is pivotal in providing precision, meaning, and structure to the answer phrase, thereby completing the question-answering part of the FAQ generation system. The outputs from this step constitute elaborate question-answer pairs that are fed along with their associated contexts to the subsequent last step.

### 3.1.6. FAQ Compilation and Ranking

The final step deals with the compilation and ranking of all question-answer pairs generated and can only be initiated after the processing of all the chunks through steps 2-5 is completed. To ensure that the generated question-answer pairs are presented to the user in an easy-to-comprehend fashion and satisfy all the requirements of FAQs, a ranking algorithm is used to organise the question-answer pairs according to their relevance to the original context. The ranking algorithm calculates a score based on two metrics that combine structural and semantic importance for each question-answer pair as follows:

- **Semantic Similarity Score:** For each question-answer pair, calculate the semantic similarity between the original context and the question-answer pair using the cosine similarity score (Gunawan et al. 2018).
- **Keyword Matching Score:** Calculate the number of keywords that exactly match between the question-answer pair and the context. Apply a penalty of (-1) keyword match if the number of keyword matches is greater than zero for every 200 characters in the question-answer pair to avoid favouring certain questions solely based on their size.

A single ranking score for each question-answer pair is calculated by taking the sum of the semantic similarity score and keyword matching score. Finally, the FAQs are presented to the end user as a complete set or as per the number of question-answer pairs required by the user by sorting them in the descending order of their ranking scores, thus completing the entire end-to-end FAQ generation system workflow.

### 3.2. Dataset Description

As described in section 2.2, a transfer learning approach was employed for achieving steps 3-5, with the primary dataset being the SQuAD (Stanford Question Answering Dataset v1.1)[6] (Rajpurkar et al. 2016), while the domain identification task was carried out using a pre-trained text classifier, as described in Section 3.1.2. The SQuAD dataset is a benchmark dataset for machine comprehension tasks in natural language processing. It consists of a collection of paragraphs paired with a set of question-answer pairs, where each question is derived from the corresponding paragraph. However, while SQuAD is a comprehensive dataset for question-answering tasks, it lacks domain-specific distinctions and answer-elaboration requirements, traits that are needed to build a comprehensive and powerful FAQ generation system. Thus, to address this lack of domain-specific datasets for question generation and the absence of specific datasets for question-answering and answer elaboration subtasks, large language model (LLM) tools were leveraged to generate and extend the available SQuAD dataset for certain subtasks.

Details of the comprehensive datasets required to train models for the subtasks of FAQ generation are summarised in Table 3 and described in detail in the following sections.

| Model | Training Dataset Description | Comments |
|---|---|---|
| Domain Identification | Pre-trained using Twitter data | - |
| Question Generation | Domain-specific datasets built by extending SQuAD with custom-generated datapoints with each dataset having at least 750 rows; separate model instances trained for each domain | Domain-specific datasets containing contexts and a list of questions; 'Questions List' column as the target column |

---

[6] https://rajpurkar.github.io/SQuAD-explorer/explore/1.1/dev/

| | | |
|---|---|---|
| Answer Keyword/Keyphrase Extraction | SQuAD + custom dataset. Total 10000 rows were utilised for training | 'Answer Phrase' column serves as the target column, 'Complete Answer' column from the custom dataset ignored |
| Answer Completion and Elaboration | Custom dataset; around 1600 rows | 'Complete Answer' column serves as the target column |

TABLE 3. DESCRIPTION OF DATASETS USED FOR TRAINING MODELS UTILISED FOR THE FAQ SYSTEM STEPS

### 3.2.1. Dataset for Question Generation

To build domain-specific datasets required to fine-tune the T5 model for question generation, questions generated from identical contexts in the SQuAD dataset were first grouped together. As the SQuAD dataset contains numerous questions covering all key points in the associated context, it was ensured that the list of questions comprehensively covered all the main ideas in the context. The domain identification model was then used to map each datapoint's (each row in the training dataset table) context to the respective domain to build 17 domain-specific datasets for question generation, the format of which is described in Table 4. Upon identifying a lack of training samples in several domains obtained from the SQuAD dataset, prompts were constructed for generative AI tools, including ChatGPT[7] and Microsoft Bing AI[8], to create custom tables and extend the domain-specific datasets to have at least 750 datapoints in each dataset. A zero-shot prompt engineering technique was used to generate the initial tabular results, and iterative prompting was utilised to add more distinct rows to the tables of each domain (see Appendix A). Finally, 17 separate instances of T5 models were fine-tuned for each domain for the question-generation task using these tabular datasets to ensure each model can generate a comprehensive list of questions containing intricate constructs and specific jargon present in that particular domain.

| Context | Questions List |
|---|---|
| In October 1810, six months after Fryderyk's birth, the family moved to Warsaw, where his father acquired a post teaching French at the Warsaw Lyceum, then housed in the Saxon Palace. Fryderyk lived with his family in the Palace grounds. The father played the flute and violin; the mother played the piano and gave lessons to boys in the boarding house that the Chopins kept. Chopin was of | What language did Frederic's father teach after they had moved to Warsaw? \| Where did Frederic live with his family while they were in Warsaw? \| What two instruments did Frederic's father play during this time? \| What instrument did Chopin's mother teach at the boarding house? |

---

[7] https://chat.openai.com/
[8] https://www.bing.com/search?q=Bing+AI&showconv=1&FORM=undexpand

| |  |
|---|---|
| slight build, and even in early childhood was prone to illnesses. | |

TABLE 4. EXAMPLE DATAPOINT (SINGLE RECORD IN THE TABLE) FROM THE 'LITERATURE' DOMAIN DATASET USED FOR TRAINING DOMAIN-SPECIFIC MODELS IN THE QUESTION-GENERATION STEP

### 3.2.2. Dataset for Answer Keyword/Keyphrase Extraction

The SQuAD dataset has a replete amount of data, which serves as the foundational knowledge for the answer keyword/keyphrase extraction task. Additionally, in order to generate high-quality, contextual and targeted answer keywords/keyphrases for contexts from various domains, a custom dataset, the structure of which is described in Table 5, was created using prompts for online generative AI tools. Similar to the process of generating training samples using generative AI tools as described in Section 3.2.1, the custom dataset was generated using a one-shot prompt engineering technique to generate the initial tabular results, with iterative prompting utilised to add more distinct rows to the table (see Appendix A). By merging the respective columns of the custom dataset (excluding the 'Complete Answer' column) with the SQuAD dataset, the dataset required to fine-tune a T5 model for the answer keyword/keyphrase extraction subtask was generated with 'Answer Phrase' as the target output.

| Context | Question | Answer Phrase | Complete Answer |
|---|---|---|---|
| Primary market research is a customised research technique that marketing professionals and businesses use to collect original information in the field. It is conducted directly by the person or organisation that stands to gain from the responses and can be performed through surveys, interviews, or focus groups. | What is primary market research? | a customised research technique to collect original information | Primary market research is a customised research technique that marketing professionals and businesses use to collect original information in the field. |

TABLE 5. EXAMPLE DATAPOINT (SINGLE RECORD IN THE TABLE) FOR THE CUSTOM-BUILT DATASET USED FOR TRAINING MODELS IN THE ANSWER KEYWORD/KEYPHRASE EXTRACTION AND ANSWER COMPLETION AND ELABORATION STEPS

### 3.2.3. Dataset for Answer Completion and Elaboration

Since the answer elaboration subtask has not been covered while designing the SQuAD dataset, the same custom dataset shown in Table 5 with the 'Complete Answer' column as the target output was the only training dataset used for training the answer completion and elaboration model of the FAQ system. This custom dataset was entirely generated using prompts for online generative AI tools, as described in Section 3.2.2.

## 3.3. Approach Merits

The modular approach described in Section 3.1 provides the following advantages over the traditional one-stroke question-answer generation workflow for the FAQ system:

- From an accuracy standpoint, each step tackles a specific subtask of the problem, making it significantly easier to fine-tune high-performing models specialised in a particular text-to-text transformation task.

- From an implementation perspective, modularisation provides greater flexibility in choosing base models, datasets, and evaluation criteria suitable for specific subtasks within the entire FAQ generation system workflow.

- From a validation point of view, a modular approach makes it easier to point out specific steps that may be underperforming by providing inaccurate or incomprehensible results. This is greatly advantageous over a single-step question-answering model as this localisation is comparatively difficult owing to the lack of intermediate results that can be cross-checked and verified.

- From a business point of view, modularisation in the system makes it possible to implement varying subscription plans based on the quality and comprehensiveness of FAQs required, ranging from minimalist to premium plan offerings.

## 4. SYSTEM INNOVATION

### 4.1. Implementation Innovation

There is a large scope for innovative features to be included in the FAQ generation system when it is provided to the end-users as a complete and easily accessible web application. This web application can not only provide a diverse array of functionalities but can also allow customisation for individual user preferences, which is essential from a business perspective. By storing historical user inputs and results obtained, users can revisit recently generated content, offering a convenient way to track and reference interactions with the system.

Through an interactive dashboard that stores document metadata, including title, generation time, number of FAQs, and other specifications, users can effectively search for the required documents and the FAQs generated. A favourites section can provide users with the ability to bookmark and save preferred content for quick and easy access, enhancing personalisation. A centralised control centre for user accounts can offer an intuitive layout for users to navigate through various features of the system seamlessly. This innovation in implementation can streamline user interactions and also add a layer of customisation, making our system more engaging and user-centric.

**4.2. Deployment Innovation**

To further enhance user interaction with the end-to-end FAQ system provided as a website, optimisations in deployment can be applied. By using the concept of multithreading, which refers to the ability of a program to run several threads of the same program at the same time, subsequently maximizing the use of available CPU (Central Processing Unit) time (Shanthi and Irudhayaraj 2009), the performance of the system can be significantly enhanced. As described in Section 3.1.1, every input text document is converted into a collection of chunks that act as independent inputs for the text-to-text transformation models downstream. Since the document is partitioned such that the text contained in every chunk, i.e., the context, is a self-reliant piece of information, it is possible to exploit parallelism by processing each chunk on a separate thread of execution for performance enhancement. Additionally, each of the four models required in steps 2-4 in Section 3.1 can be loaded on separate threads. This makes it possible for every model to remain engaged in its dedicated task while also reducing idle time, resulting in vast performance optimisations.

## 5. RESULTS AND DISCUSSION

In order to comprehensively analyse the capabilities of the FAQ generation system, multiple tests encompassing several documents containing content from each of the 17 domains mentioned in Table 2 were fed to the system provided as a web application. The input test documents were prepared by providing customised prompts to generative AI systems like Bing AI and ChatGPT to generate varied paragraphs based on the required domain.

For each test, a PDF document containing text of size up to 2000 words relevant to a particular domain along with a random number signifying the number of FAQs to be generated was provided as input, and FAQs ranked in decreasing order of relevance were obtained as the output, as shown in Figure 4, to generate result documents. If the number of FAQs specified was greater than the number of question-answer pairs generated by the system, a warning was also provided by the system that further generation from the same content would lead to repeated and low-quality FAQs. As automated testing methodologies would fail to identify and judge the intricate criteria and qualities needed for good-quality FAQs and as no benchmarks for the separate NLP task of FAQ generation exist, a qualitative review methodology was chosen to capture and analyse the performance of the system. Accordingly, independent human judges carried out a qualitative, subjective review of the outputs in each result document.

Each result document was reviewed by exactly four human judges. A total of sixteen human judges identified to be highly proficient in English through standardised language test scores took part in the review process. The review process was carried out by circulating the result documents along with the domain the content belongs to, accompanied by a form to provide scores for each document based on five test questions. Detailed grading instructions and the significance of each test question were provided to each reviewer who consented to the judging process (see Appendix B). Reviewers were asked to provide scores out of 10 for the output FAQs, considered as a collection, based on the following five test questions, with their significance explained in brackets:

(1) Is the set of questions syntactically well-formed? (To evaluate the questions based on readability and grammatical construction)
(2) Are answers to questions syntactically well-formed? (To evaluate the answers based on readability and grammatical construction)
(3) Are the questions relevant to the input passage? (To evaluate the relevance of questions by checking if the questions cover all topics, subtopics and key ideas in the input passage)
(4) Are answers relevant to the questions, and do they answer satisfactorily? (To evaluate the relevance of answers to the corresponding questions by checking if the answers cover all aspects put forth by the questions and include all relevant points from the input passage)
(5) Does the set of questions comprehensively cover the passage? (To evaluate the ability of the set of FAQs to accurately and comprehensively represent all the information in the original text)

---

**Input:**
In the dynamic realm of modern fashion, trends emerge and evolve with remarkable speed, reflecting the ever-changing tastes, lifestyles, and cultural influences of contemporary society. Fashion serves as a powerful form of self-expression, allowing individuals to communicate their identities, values, and aspirations through clothing, accessories, and personal style. Sustainability and ethical practices have emerged as critical considerations within the fashion industry, prompting calls for transparency, accountability, and responsible consumption. From eco-friendly materials and supply chain transparency to fair labour practices and circular business models, fashion brands and consumers alike seek to minimize environmental impact and promote social equity throughout the fashion lifecycle. Fashion activism and cultural movements further challenge industry norms and stereotypes, advocating for diversity, inclusion, and representation across body types, gender identities, and cultural backgrounds. As fashion continues to evolve in response to societal shifts and technological advancements, it remains a vibrant expression of creativity, identity, and cultural zeitgeist in the modern era.
**Question Count:** 2

**Output:**
**"question":** How does technology influence modern fashion?
**"answer":** Technology revolutionizes design, manufacturing, and retail processes, democratizing trends and fostering global fashion communities online.

**"question":** What drives sustainability efforts in the fashion industry?
**"answer":** Sustainability prompts brands and consumers to prioritize eco-friendly materials and ethical practices, minimizing environmental impact and promoting social equity in fashion.

---

FIGURE 4. EXAMPLE INPUT AND OUTPUT FROM A TEST DOCUMENT BELONGING TO THE DOMAIN OF 'CELEBRITY AND FASHION'.[9]

---

[9] A few select samples of test inputs and outputs from all domains can be found at https://github.com/Sahil-R-Kale/faq-result-samples.git

The automated FAQ generation system produces a collection of question-answer pairs as output. Since FAQs are characterised as a complete and coherent set, their evaluation also requires considering them as a collection. Hence, by evaluating the output FAQs from the system in each result document as a collection, a better assessment of the system's overall performance was achieved, rather than by grading each question-answer pair individually. Scores obtained from the four judges for all result documents in a given domain were averaged and rounded off to the nearest whole number for each of the five test questions described above to measure the system performance for a particular domain. These results are presented in Table 6.

Also, in order to measure the inter-agreement among the reviewers while assigning scores, the standard deviation of average scores given by each reviewer for all documents in a given domain was calculated. Table 7 thus provides an idea of the agreement among reviewers while assigning scores to documents in a given domain by showing these standard deviation values for each of the 5 test questions.

| Domain | No. of test documents (with no. of question-answer pairs generated for each document in brackets) | Test Question 1 | Test Question 2 | Test Question 3 | Test Question 4 | Test Question 5 |
|---|---|---|---|---|---|---|
| Arts and Culture | 5 (4,5,3,6,7,2) | 9 | 7 | 9 | 8 | 9 |
| Business and Entrepreneurs | 3 (7,3,4) | 8 | 8 | 4 | 9 | 5 |
| Celebrity and Fashion | 4 (2,7,6,4) | 9 | 9 | 8 | 8 | 10 |
| Diaries and Daily Life | 3 (3,3,4) | 7 | 7 | 6 | 6 | 3 |
| Family and Relationships | 5 (7,2,2,4,3) | 10 | 9 | 9 | 8 | 8 |
| Film, TV and Video | 6 (4,6,4,3,2,6) | 6 | 7 | 8 | 6 | 9 |
| Fitness and Health | 4 (2,5,3,4) | 9 | 10 | 10 | 9 | 8 |
| Food and Dining | 5 (3,5,6,4) | 8 | 7 | 8 | 7 | 5 |
| Gaming | 3 (7,3,2) | 8 | 4 | 7 | 3 | 6 |
| Learning and Educational | 5 (2,4,4,3,6) | 9 | 8 | 9 | 8 | 9 |
| Literature | 4 (5,4,3,4) | 8 | 9 | 4 | 7 | 4 |
| Music | 3 (6,3,4) | 7 | 8 | 7 | 9 | 7 |
| News and Social Concern | 6 (4,3,2,6,7,3) | 9 | 10 | 10 | 10 | 9 |
| Science and Technology | 5 (6,3,4,5,2) | 8 | 9 | 9 | 8 | 8 |
| Sports | 6 (5,3,4,6,2) | 9 | 8 | 8 | 9 | 10 |
| Travel and Adventure | 4 (4,4,3,5) | 7 | 9 | 8 | 9 | 7 |
| Youth and Student Life | 3 (3,3,4) | 7 | 7 | 5 | 6 | 4 |
| **Overall Average** | **4.4** | **8.1** | **8.0** | **7.6** | **7.7** | **7.1** |

TABLE 6. AVERAGE SCORES OF THE QUALITATIVE SUBJECTIVE REVIEW CARRIED OUT ON MULTIPLE OUTPUTS OBTAINED FROM THE FAQ GENERATION SYSTEM GIVEN BY INDEPENDENT HUMAN JUDGES ORGANISED BY DOMAIN AND TEST QUESTION

| Domain | No. of test documents | Standard Deviation for | | | | |
|---|---|---|---|---|---|---|
| | | **Test Question 1** | **Test Question 2** | **Test Question 3** | **Test Question 4** | **Test Question 5** |
| Arts and Culture | 5 | 0.49 | 0.43 | 0.00 | 0.49 | 0.80 |
| Business and Entrepreneurs | 3 | 0.47 | 0.47 | 1.24 | 1.11 | 1.24 |

| Domain | | | | | | |
|---|---|---|---|---|---|---|
| Celebrity and Fashion | 4 | 0.50 | 0.43 | 1.47 | 0.83 | 0.50 |
| Diaries and Daily Life | 3 | 1.11 | 0.82 | 1.11 | 0.47 | 0.82 |
| Family and Relationships | 5 | 0.40 | 0.80 | 0.49 | 1.02 | 0.80 |
| Film, TV and Video | 6 | 0.63 | 0.47 | 0.37 | 1.70 | 0.47 |
| Fitness and Health | 4 | 0.50 | 0.00 | 0.83 | 0.43 | 0.83 |
| Food and Dining | 5 | 0.75 | 0.80 | 0.75 | 0.80 | 1.41 |
| Gaming | 3 | 0.47 | 0.83 | 1.63 | 1.11 | 0.83 |
| Learning and Educational | 5 | 0.49 | 0.75 | 0.00 | 0.75 | 1.02 |
| Literature | 4 | 0.83 | 0.50 | 1.47 | 0.50 | 1.47 |
| Music | 3 | 0.00 | 0.47 | 1.11 | 0.83 | 1.11 |
| News and Social Concern | 6 | 0.37 | 0.50 | 0.37 | 0.37 | 0.50 |
| Science and Technology | 5 | 0.63 | 0.49 | 0.00 | 0.49 | 0.80 |
| Sports | 6 | 0.37 | 0.50 | 1.21 | 0.80 | 0.37 |
| Travel and Adventure | 4 | 0.70 | 0.83 | 0.43 | 0.83 | 0.83 |
| Youth and Student Life | 3 | 0.47 | 0.47 | 1.11 | 1.63 | 1.11 |
| **Overall Average** | **4.4** | **8.1** | **8.0** | **7.6** | **7.7** | **7.1** |

TABLE 7. STANDARD DEVIATION AMONG AVERAGE SCORES ASSIGNED BY INDEPENDENT HUMAN JUDGES TO RESULT DOCUMENTS IN A PARTICULAR DOMAIN FOR EACH TEST QUESTION

The findings and inferences from the scores obtained after the qualitative review can be presented as follows:

- High overall average scores, along with low deviation among reviewers, are seen in the first two test questions for all domains in Table 6 and Table 7, which grade the syntactic quality of questions and answers. The modular nature of the system and the powerful text-to-text transformation models used thus allowed the system to successfully generate well-formed, grammatically correct and easy-to-read questions and answers.
- Domains having a larger size of training data performed significantly better in generating both syntactically correct and semantically relevant questions and answers, as well as FAQs that comprehensively cover the passage, as expected. It can thus be inferred that the performance of the system for domains having weaker scores in Table 6 can be improved by increasing the size of the training datasets.
- Despite the process of judging the relevance of questions and answers and the coverage of the FAQ sets being highly subjective, domains having high average scores have significantly lower standard deviation values, showing confidence among the reviewers while assigning positive scores. This clearly showcases the system's ability to generate relevant and comprehensive FAQs when trained with sufficient domain-specific data.
- Domains that are traditionally considered to have a large number of specific terminologies, jargon, and special constructs, such as 'Arts and Culture', 'Science and Technology', and 'Sports' have high scores with high agreement among reviewers for all test questions, highlighting the increase in performance obtained by training domain-specific models.
- Finally, the highest average scores and minimal deviation among scores for the 'News and Social Concern' domain, which majorly contains generalised articles and text with non-technical wordings, showcase the ability of the system to deal with generic content effectively without needing domain-specific additions.

## 6. CONCLUSION AND FUTURE SCOPE

Through this paper, we propose and showcase the effectiveness of building a systematic end-to-end process pipeline that generates high-quality, information-rich, and coherent FAQs for any given text document. We demonstrate state-of-the-art qualitative results for the FAQ generation task obtained from the system. Special attention has been paid to bringing completeness to the FAQ Generation process and treating it as a complete NLP task by bridging the gaps in contemporary question-answering approaches. All text-to-text transformation models are trained to leverage domain-specific vocabulary, and extra care has been taken to present the generated FAQs in a human-readable format ranked by relevance. Additionally, custom datasets are generated where publicly available datasets fail to capture the required features for the training dataset. We also utilise self-curated algorithms to parse the inputs and rank the outputs. We believe that the approach proposed and the NLP artefacts, including dataset formats, algorithms, and architectures conceptualised through this paper, will provide valuable pointers to NLP practitioners and the research community as a whole for further exploration.

Further enhancements to the system include providing end-users the ability to modify and customise the generated FAQs through a user-friendly interface and re-generating FAQs based on the modifications made. By incorporating sentiment analysis into the system, FAQs whose language reflects user moods, including chirpiness, frustration, or curtness, can also be generated, fostering a more positive and user-centric content management experience. Information customised according to user expertise and experience levels can also be provided by categorising FAQs into different difficulty tiers. Finally, integration with external databases and knowledge repositories can help the system in providing more advanced question types with detailed, in-depth answers.

## ACKNOWLEDGEMENTS

The research for this paper was carried out for the 'ROME - Automated FAQ Engine' project initiated and funded by Stride.ai R&D Pvt Ltd, Bengaluru, India. We gratefully acknowledge the support of Mr. Vijaykant Nadadur, Co-Founder & CEO, Stride.ai and Mr. Raghav Khamar, Senior Technical Project Manager, Stride.ai, in the core ideation and development of this project. Finally, we also acknowledge the work of the entire team at Stride.ai for their continuous and unwavering support throughout the development of this system.

## REFERENCES

Alyafeai, Zaid, Maged Saeed AlShaibani, and Irfan Ahmad. 2020. "A Survey on Transfer Learning in Natural Language Processing". *arXiv:2007.04239 [cs.CL]*. http://arxiv.org/abs/2007.04239.

Antypas, Dimosthenis, Asahi Ushio, Jose Camacho-Collados, Vitor Silva, Leonardo Neves, and Francesco Barbieri. 2022. "Twitter Topic Classification". In *Proceedings of the 29th International Conference on Computational Linguistics*, edited by Nicoletta Calzolari, Chu-Ren Huang, Hansaem Kim, James Pustejovsky, Leo Wanner, Key-Sun Choi, Pum-Mo Ryu, et al., 3386–3400. https://aclanthology.org/2022.coling-1.299.

Chan, Ying-Hong, and Yao-Chung Fan. 2019. "A Recurrent BERT-Based Model for Question Generation". In *Proceedings of the 2nd Workshop on Machine Reading for Question Answering*, edited by Adam Fisch, Alon Talmor, Robin Jia, Minjoon Seo, Eunsol Choi, and Danqi Chen, 154–62. https://doi.org/10.18653/v1/D19-5821.

Das, Bidyut, Mukta Majumder, Santanu Phadikar, and Arif Ahmed Sekh. 2021. "Automatic Question Generation and Answer Assessment: A Survey." *Research and Practice in Technology Enhanced Learning* 16 (1). https://doi.org/10.1186/s41039-021-00151-1

Devlin, Jacob, Chang, Ming-Wei, Lee, Kenton, and Toutanova, Kristina. 2019. "BERT: Pre-training of Deep Bidirectional Transformers for Language Understanding". In *Proceedings of the 2019 Conference of the North American Chapter of the Association for Computational Linguistics*: *Human Language Technologies*, 4171–4186. https://aclanthology.org/N19-1423/

Du, Xinya and Cardie, Claire. 2018. "Harvesting Paragraph-level Question-Answer Pairs from Wikipedia". In *Proceedings of the 56th Annual Meeting of the Association for Computational Linguistics*, 1907–1917. https://aclanthology.org/P18-1177

Du, Xinya, Shao, Junru and Cardie, Claire. 2017. "Learning to Ask: Neural Question Generation for Reading Comprehension. In *Proceedings of the 55th Annual Meeting of the Association for Computational Linguistics*, 1342–1352. https://aclanthology.org/P17-1123

Gunawan, Dani, Sembiring, C.A., and Budiman, Mohammed. 2018. "The Implementation of Cosine Similarity to Calculate Text Relevance between Two Documents". *Journal of Physics: Conference Series*, *978, 012120*. https://doi.org/10.1088/1742-6596/978/1/012120

Hu, Wenpeng, Liu, Bing, Ma, Jinwen, Zhao, Dongyan, and Yan, Rui. 2018. "Aspect-based Question Generation. 6th International Conference on Learning Representations". In *Workshop Track Proceedings of ICLR 2018*, April 30 - May 3. https://openreview.net/forum?id=rkRR1ynIf

Joshi, Mandar, Choi, Eunsol, Weld, Daniel, and Zettlemoyer, Luke. 2017. "TriviaQA: A Large Scale Distantly Supervised Challenge Dataset for Reading Comprehension". In *Proceedings of the 55th Annual Meeting of the Association for Computational Linguistics*, 1601–1611. https://aclanthology.org/P17-1147/

Kay, Anthony. 2007. "Tesseract: an open-source optical character recognition engine". *Linux Journal*. 159: 2. https://dl.acm.org/doi/10.5555/1288165.1288167

Kočiský, Tomáš, Schwarz, Jonathan, Blunsom, Phil, Dyer, Chris, M. Hermann, Karl, Melis, Gábor, and Grefenstette, Edward. 2018. "The NarrativeQA Reading Comprehension Challenge". *Transactions of the Association for Computational Linguistics*, 6:317–328. https://aclanthology.org/Q18-1023/

Kumar, Vishwajeet, Chaki, Raktim, Talluri, Sai T., Ramakrishnan, Ganesh, Li, Yuan-Fang, and Haffari, Gholamreza. 2019. "Question Generation from Paragraphs: A Tale of Two Hierarchical Models". *arXiv: 1911.03407 [cs.CL]*. https://arxiv.org/abs/1911.03407

Kunichika, Hidenobu, Katayama, Tomoki, Hirashima, Tsukasa, and Takeuchi, Akira. 2004. "Automated question generation methods for intelligent English learning systems and its evaluation". In *Proceedings of ICCE, 2004.*

Liu, Bang, Wei, Haojie, Niu, Di, Chen, Haolan, and He, Yancheng. 2020. "Asking Questions the Human Way: Scalable Question-Answer Generation from Text Corpus". In *Proceedings of The Web Conference 2020*, 2032–2043. https://doi.org/10.1145/3366423.3380270

Lu, Chao.-Yi, and Lu, Sin-En. 2021. "A Survey of Approaches to Automatic Question Generation: From 2019 to Early 2021". In *Proceedings of the 33rd Conference on Computational Linguistics and Speech Processing (ROCLING 2021)*, 151–162. https://aclanthology.org/2021.rocling-1.21

Raazaghi, Fatemah. 2015. "Auto-FAQ-Gen: Automatic Frequently Asked Questions Generation". In *Advances in Artificial Intelligence*, edited by Denilson Barbosa and Evangelos Milios, 334–37, Springer International Publishing.

Raffel, Colin, Shazeer, Noam, Roberts, Adam, Lee, Katherine, Narang, Sharan, Matena, Michael, Zhou, Yanqi, Li, Wei, and Liu, Peter J. 2023. "Exploring the Limits of Transfer Learning with a Unified Text-to-Text Transformer". *Journal of Machine Learning Research* 21: 1-67. https://jmlr.org/papers/volume21/20-074/20-074.pdf

Rajpurkar, Pranav, Zhang, Jian, Lopyrev, Konstantin, and Liang, Percy. 2016. "SQuAD: 100,000+ Questions for Machine Comprehension of Text". In *Proceedings of the 2016 Conference on Empirical Methods in Natural Language Processing*, 2383–2392. https://aclanthology.org/D16-1264/

Roemmele, Melissa, Sidhpura, Deep, DeNeefe, Steve, and Tsou, Ling. 2021. "AnswerQuest: A System for Generating Question-Answer Items from Multi-Paragraph Documents". In *Proceedings of the 16th Conference of the European Chapter of the Association for Computational Linguistics: System Demonstrations*, 40–52. https://aclanthology.org/2021.eacl-demos.6/

Ruder, Sebastian, E. Peters, Matthew, Swayamdipta, Swabha and Wolf, Thomas. 2019. "Transfer Learning in Natural Language Processing". In *Proceedings of the 2019 Conference of the North American Chapter of the Association for Computational Linguistics: Tutorials*, 15–18. https://aclanthology.org/N19-5004/

Shanthi, M., and Irudhayaraj, Anthony A. 2009. "Multithreading-An Efficient Technique for Enhancing Application Performance". *International Journal of Recent Trends in Engineering,* 2 (4).

Shen, Sheng, Yaliang Li, Nan Du, Xian Wu, Yusheng Xie, Shen Ge, Tao Yang, Kai Wang, Xingzheng Liang, and Wei Fan. 2020. "On the Generation of Medical Question-Answer Pairs". In *Proceedings of the AAAI Conference on Artificial Intelligence* 34 (05):8822-29. https://doi.org/10.1609/aaai.v34i05.6410.

Trischler, Adam, Wang, Tong, Yuan, Xingdi, Harris, Justin, Sordoni, Alessandro, Bachman, Philip, and Suleman, Kaheer. 2017. "NewsQA: A Machine Comprehension Dataset". In *Proceedings of the 2nd Workshop on Representation Learning for NLP*, 191–200. https://aclanthology.org/W17-2623/

Zhang, Shiyue and Bansal, Mohit. 2019. "Addressing Semantic Drift in Question Generation for Semi-Supervised Question Answering". In *Proceedings of the 2019 Conference on Empirical Methods in Natural Language Processing and the 9th International Joint Conference on Natural Language Processing (EMNLP-IJCNLP)*, 2495–2509. https://aclanthology.org/D19-1253/

Zhang, tong, Liu, Yong, Li, Boyang, Zeng, Zhiwei, Wang, Pengwei, You, Yuan, Miao Chunyan, and Cui, Lizhen. 2022. "History-Aware Hierarchical Transformer for Multi-session Open-domain Dialogue System". In *Findings of the Association for Computational Linguistics: EMNLP 2022*, 3395–3407. https://aclanthology.org/2022.findings-emnlp.247/

Zhao, Yao, Ni, Xiaochuan, Ding, Yuanyuan, and Ke, Qifa. 2018. "Paragraph-level Neural Question Generation with Maxout Pointer and Gated Self-attention Networks". In *Proceedings of the 2018 Conference on Empirical Methods in Natural Language Processing*, 3901–3910. https://aclanthology.org/D18-1424/

Zhou, Qingyu, Yang, Nan, Wei, Furu, Tan, Chuanqi, Bao, Hangbo, & Zhou, Ming. 2017. "Neural Question Generation from Text: A Preliminary Study". *arXiv:1704.01792 [cs.CL]*. https://arxiv.org/abs/1704.01792

# APPENDIX A

**PROMPT TEMPLATE TO GENERATE DATASETS FOR QUESTION GENERATION**

The following template was used as a prompt to generate the initial tabular result for each domain-specific dataset for question generation:

> Hello! I need your help with an NLP task. I want to generate a dataset in the following format:
>
> Column 1: "Context", which is a paragraph of 250 words related only to the domain of <required domain>.
>
> Column 2: "Question List", a list of questions whose answers are present in the paragraph in column 1. Make sure the questions cover every key point in the paragraph. Separate the individual questions with a | character.
>
> Please do this nicely for me. Draw in your table to display the results.

The following prompt was then iteratively used to generate subsequent results for the same domain:

> Great work! Now, I want you to continue this same process with the same format. Could you please give me more distinct samples in the same domain of <required domain>? Ensure that the context column has 250 words. Draw in your table to display the results.

**PROMPT USED TO GENERATE THE CUSTOM DATASET FOR ANSWER KEYWORD/KEYPHRASE EXTRACTION AND ANSWER COMPLETION**

The following prompt was used to generate the initial tabular result for the custom dataset needed for both answer keyword/keyphrase extraction and answer completion:

> Hello! I need your help with an NLP task. I want to generate a dataset having the following format:
>
> Context, Question, Answer Phrase or Keywords, Complete Answer
>
> For example:
>
> Context: "The American Civil War, which lasted from 1861 to 1865, was a major conflict that occurred in the United States. It was primarily a result of tensions between the Northern states, which were industrialised and anti-slavery, and the Southern states, which were agrarian and pro-slavery. One of the key events that led to the war was the election of Abraham Lincoln as the 16th President of the United States in 1860, which prompted several Southern states to secede from the Union. The conflict saw significant battles, including Gettysburg and Antietam, and it ended with the surrender of the Confederacy in 1865."

Question: What was the main cause of the American Civil War?

Answer Phrase or Keywords: tensions between Northern and Southern states

Complete Answer: The main cause of the American Civil War was the tensions between Northern and Southern states in the United States.

Diversify the domains and topics of the context as much as possible. Ensure that the context column has 250 words. Please do this nicely for me. Draw the results in your table.

The following prompt was then iteratively used to generate subsequent results:

Great work! Now, I want you to continue this same process with the same format. Could you please give me more distinct samples? Ensure that the context column has 250 words. Draw in your table to display the results.

## APPENDIX B

The following grading instructions were provided to each reviewer, accompanied by a form, result documents and the domain of each result document:

Dear Reviewer,

Thank you for participating in the review process for our automated FAQ generation system. Your feedback is invaluable in ensuring the quality and accuracy of our system's output. You will have received the result documents grouped by the domain they belong to in a zip file. Below are detailed grading instructions along with some principles to ensure fairness and accuracy in your evaluations:

- Fairness: Treat each document impartially and independently, focusing solely on the quality and relevance of the generated FAQs.

- Accuracy: Ensure your evaluations are based on objective criteria, avoiding biases or personal opinions.

- Thoroughness: Review each document comprehensively, considering all aspects of the generated FAQs and their alignment with the input passage.

Since FAQs are characterised as a complete and coherent set, their evaluation also requires considering them as a collection. There are five test questions as follows that you must answer by giving a score out of 10 for each document:

(1) Is the set of questions syntactically well-formed? (Evaluate the questions based on readability and grammatical construction)

(2) Are answers to questions syntactically well-formed? (Evaluate the answers based on readability and grammatical construction)

(3) Are the questions relevant to the input passage? (Evaluate the relevance of questions by checking if the questions cover all topics, subtopics and key ideas in the input passage)

(4) Are answers relevant to the questions, and do they answer satisfactorily? (Evaluate the relevance of answers to the corresponding questions by checking if the answers cover all aspects put forth by the questions and include all relevant points from the input passage)

(5) Does the set of questions comprehensively cover the passage? (Evaluate the ability of the set of FAQs to accurately and comprehensively represent all the information in the original text)

Please review each document and fill in the name of the document, the domain it belongs to, and your score for each test question out of 10 in the attached form. Please repeat the process for every result document.

Optionally, feel free to justify your scores with reference to the provided FAQs and input passage, citing examples where necessary in the "Notes" section of the form. Your contribution to this review process is highly appreciated.